\documentclass[11pt]{article}

\usepackage{hyperref}

\usepackage{natbib}

\usepackage{setspace}
\usepackage{geometry}
\usepackage[section]{placeins}
\usepackage{graphicx}
\usepackage{xcolor}
\usepackage{titlesec}
\usepackage[page]{appendix}
\usepackage{enumerate}

\usepackage{subfiles}

\usepackage{lscape}
\usepackage{booktabs}
\usepackage{rotating}
\usepackage{multirow}
\usepackage{longtable}
\usepackage{caption}
\usepackage{subcaption}
\usepackage{float}
\usepackage{tabularx}
\usepackage{ragged2e}
\newcolumntype{Y}{>{\RaggedRight\arraybackslash}X}
\usepackage{pdflscape}
\usepackage{afterpage}

\usepackage{amsmath}
\usepackage{amssymb}
\usepackage{amsthm}
\usepackage{mathtools}
\usepackage{dsfont}

\usepackage[ruled,vlined]{algorithm2e}

\SetCommentSty{algcommstyle}

\usepackage{tikz}
\usetikzlibrary{bayesnet}

\usepackage{textcomp}
\usepackage{sourcecodepro}
\usepackage{listings}
\definecolor{commentgrey}{gray}{0.45}
\definecolor{backgray}{gray}{0.96}
\DeclareMathOperator*{\CinP}{\overset{P}\rightarrow}

\DeclareMathOperator*{\argmin}{arg\,min}

\newcommand*{\cond}{\;\ifnum\currentgrouptype=16 \middle\fi|\;}

\newcommand*{\ttilde}{{\raise.17ex\hbox{$\scriptstyle\sim$}}}

\makeatletter
\newsavebox{\mybox}\newsavebox{\mysim}
\newcommand*{\distas}[1]{%
  \savebox{\mybox}{\hbox{\kern3pt$\scriptstyle#1$\kern3pt}}%
  \savebox{\mysim}{\hbox{$\sim$}}%
  \mathbin{\overset{#1}{\kern\z@\resizebox{\wd\mybox}{\ht\mysim}{$\sim$}}}%
}
\makeatother

\makeatletter
\def\moverlay{\mathpalette\mov@rlay}
\def\mov@rlay#1#2{\leavevmode\vtop{%
   \baselineskip\z@skip \lineskiplimit-\maxdimen
   \ialign{\hfil$\m@th#1##$\hfil\cr#2\crcr}}}
\newcommand*{\charfusion}[3][\mathord]{
  #1{\ifx#1\mathop\vphantom{#2}\fi\mathpalette\mov@rlay{#2\cr#3}}
  \ifx#1\mathop\expandafter\displaylimits\fi}
\makeatother

\newtheorem{theorem}{Theorem}[section]
\newtheorem{theorem*}{Theorem}
\newtheorem{corollary}[theorem]{Corollary}%[section]
\newtheorem{corollary*}[theorem*]{Corollary}
\newtheorem{proposition}[theorem]{Proposition}%[section]
\newtheorem{proposition*}[theorem*]{Proposition}
\newtheorem{lemma}[theorem]{Lemma}%[section]
\newtheorem{lemma*}[theorem*]{Lemma}

\newtheorem{property}{Property}
\newtheorem{condition}{Condition}
\newtheorem{reg_condition}{Regularity Condition}

\theoremstyle{definition}
\newtheorem{definition}{Definition}[section]
\newtheorem{definition*}{Definition}

\newtheorem*{remark}{Remark}

\newtheoremstyle{algodesc}{}{}{}{}{\bfseries}{.}{ }{}%
\theoremstyle{algodesc}

\begin{document}

\title{Not All are Made Equal: Consistency of Weighted Averaging Estimators Under Active Learning}
\author{
    Jack Goetz, Ambuj Tewari\\
    University of Michigan
}
\date{\today}

\maketitle

\begin{abstract}
  Active learning seeks to build the best possible model with a budget of labelled data by sequentially selecting the next point to label. However the training set is no longer \textit{iid}, violating the conditions required by existing consistency results. Inspired by the success of Stone's Theorem we aim to regain consistency for weighted averaging estimators under active learning. Based on ideas in \citet{dasgupta2012consistency}, our approach is to enforce a small amount of random sampling by running an augmented version of the underlying active learning algorithm. We generalize Stone's Theorem in the noise free setting, proving consistency for well known classifiers such as $k$-NN, histogram and kernel estimators under conditions which mirror classical results. However in the presence of noise we can no longer deal with these estimators in a unified manner; for some satisfying this condition also guarantees sufficiency in the noisy case, while for others we can achieve near perfect inconsistency while this condition holds. Finally we provide conditions for consistency in the presence of noise, which give insight into why these estimators can behave so differently under the combination of noise and active learning. 
\end{abstract}

\vspace{-8pt}

\section{INTRODUCTION} \label{sect:intro}

Active learning results in training data which is neither independent, nor from the same distribution on our covariates as the test data (which we assume we have no control over and which is drawn \textit{iid} from some underlying joint distribution). Thus even if our classification algorithm is well studied, standard results on consistency of that classifier, arguably the minimal requirement for a good method, no longer apply. The loss of consistency is of practical concern as even popular active learning algorithms can induce inconsistency \citep{dasgupta2011two}. Can we recover this lost consistency? 

We begin to answer this question by focusing on weighted averaging binary classifiers, of which $k$-NN, histogram and kernel estimators \citep{devroye2013probabilistic} are the classic examples. Under \textit{iid} assumptions consistency of these is largely covered by the celebrated Stone's Theorem \citep{stone1977consistent}, and our goal is to generalize these results to an actively selected training set. However it is clear that if our active learning method can be completely arbitrary, there is not much hope of obtaining consistency. Adapting a requirement in  \citet{dasgupta2012consistency}, we begin by introducing a method to \textit{augment} any existing active learning algorithm, which only influences the sampling policy a vanishing fraction of the time. 

In the noiseless setting this augmentation is sufficient, and consistency of the above classical estimators is proven using a technical condition. However in the presence of noise the behaviour of these classical estimators diverges sharply; for histogram estimators satisfying this condition guarantees consistency even with noise, whereas for $k$-nn we provide a counterexample where the condition is satisfied, but we achieve maximal Risk. Finally we will provide additional conditions which provide consistency under noise, and which illustrate the differences between histogram and $k$-nn which lead to starkly different behaviour. 

%We derive a condition on our classifier such that in the noiseless setting this augmentation is sufficient to ensure consistency. However in the presence of noise such an augmentation is not always sufficient. We provide conditions for consistency in the presence of noise, which can be satisfied by some classical weighted averaging estimators but not others. 

The structure of our paper is as follows:

\begin{enumerate}
    \item Give a natural augmentation to any sequential active learning algorithms (Algorithm \ref{alg:generic}).
    \item Proving that in the noiseless setting and under this augmentation, histogram and $k$-nn are consistent (\ref{prop:noiseless_all}). These are proved by providing a sufficient condition (Condition \ref{property:noiseless}) for consistency for any weighted averaging estimator (Theorem \ref{thrm:noiseless_case}).
    \item Showing the histogram estimator is still consistent under this condition even in the noisy setting (\ref{prop:histogram}).
    \item Providing a counterexample in the noisy setting where $k$-nn satisfies our condition, but achieves the largest Risk possible (Theorem \ref{thrm:1nn}).
    \item Provide further conditions (Condition \ref{property:bounded_sup}) which are sufficient for consistency in the noisy setting, which show why histogram is sufficient but $k$-nn is not (Theorem \ref{thrm:bounded_sup_consistency}).
\end{enumerate}

\section{SETTING AND BACKGROUND} \label{sect:setting}

Our positive results will be in the query synthesis setting, where as our negative result will be in the pool setting (which is the setting in which the negative result is more challenging). Our setup will be fairly standard for active learning \citep{settles2012active}. In the query synthesis setting the active learning algorithm can select any point within the support. In the pool setting the algorithm will select $n$ data points to label out of a pool of $m_n$ data points, where the size of our initial pool depends on how many labelled points we will select. Let $D_n = \{(X_i, Y_i)\}_{i=1}^{m_n}$ be our pool with known covariates $X_i \in \mathcal{X} \subset \mathcal{R}^d$ and hidden labels $Y_i \in \{0,1\}$, where $(X_i, Y_i) \stackrel{iid}{\sim} P_{X,Y} = P_{Y|X} P_X$, $f(x) = P(Y = 1 | X = x)$ and with Bayes classifier $f^*(x) = \mathbf{1}_{f(x) > 1/2}$. We will assume that $\mathcal{X}$ is a bounded subset of $\mathcal{R}^d$, however if this does not hold then many of our results can be applied on a sphere centered at the origin with all but an arbitrary $\epsilon$ of the probability mass to extend the results beyond bounded $\mathcal{X}$. Additionally let $D_n(X)$ and $D_n(Y)$ denote just the $X$ and $Y$ of the pool respectively. Note that the pool setting is slightly different from the setup in \citet{hanneke2014theory}, as our setting assumes $m_n < \infty$ while theirs assumes $m_n = \infty \: \forall \: n$.

The algorithm will create a labelled subset $S_n$ with the goal of minimizing the risk $E \mathbf{1}_{f_{n}(X, S_n) \neq Y}$. The notation $f_n(x, S_n)$ indicates the prediction given at point $x$ when trained on the labelled data $S_n$ (with $S_n(X), S_n(Y)$ as just the covariates and labels). We use lower case letter $x$ to denote non-random quantities and upper case $X$ to denote random ones. We will use \textit{passive sampling} to refer to sampling according to the marginal $P_X$. In the pool setting given $S_n$, let $S_n^c$ be the remaining $m_n - n$ unlabelled data points, with $\emptyset^c = D_n(X)$ (so it's not exactly a true complement operator but has a similar flavor). Our (potentially randomized) active learning algorithm selecting the $i^{th}$ point will be $A: S_{i-1} \rightarrow supp(\mathcal{X})$ in the query synthesis setting and $A: S_{i-1} \times S_{i-1}^c \rightarrow S_{i-1}^c$ in the pool setting. Technically $S_n$ is a multiset and so can contain identical 2-tuples $(X_i, Y_i)$. 

We will focus on weighted averaging estimators for classification \citep{devroye2013probabilistic}, where the estimators take the following form (where $W_{ni}(x) = W_{ni}(x, S_n(X))$)

\begin{equation*}
\begin{aligned}
    &f_n(x, S_n) = 
    \begin{cases}
    0 & \text{ if } \sum\limits_{(X_i, Y_i) \in S_n} Y_i W_{ni}(x) \leq \frac{1}{2}\\
    1 & \text{ otherwise}\\
    \end{cases}\\
\end{aligned}
\end{equation*}

We will make the following assumptions about the structure of our functions $W_{ni}(x)$.

\begin{equation*}
\begin{aligned}
    W_{ni}(x) \geq 0, \quad \sum W_{ni}(x) \leq 1\\
\end{aligned}
\end{equation*}

% \begin{enumerate}
%     \item $W_{ni}(x) \geq 0$.
%     \item $\sum W_{ni}(x) \leq 1$.
% \end{enumerate}

% We are only interested in the performance of our estimator trained on all $n$ data points, and will assume that once the $n$ labelled data points are selected, the estimator has no knowledge of the other unlabelled data.

% We will use the standard $k_n$-NN estimator for classification \citep{devroye2013probabilistic}, where the number of points we take the mode over ($k_n$) is allowed to increase as the number of (labelled) samples increases. Ties will be broken with a fair coin flip. We will only be interested in the performance of our estimator trained on all $n$ data points, and will assume that once the $n$ labelled data points are selected, the estimator has no knowledge of the other unlabelled data. Of particular interest to us is the famous theorem due to \cite{stone1977consistent} which shows that if $k_n \rightarrow \infty, \: \frac{k_n}{n} \rightarrow 0$ then $k_n$-NN is universally consistent.

The inconsistency introduced during active learning is well documented, where even in the one dimensional case popular and intuitive algorithms can be inconsistent in non-pathological examples \citep{dasgupta2011two}. A recent study \citep{loog2016empirical} showed that while most active learning methods examined performed well on many data sets, they also had data sets on which they do not appear to be converging to the performance of random sampling. Our work extends that of \cite{dasgupta2012consistency}, which studied consistent active learning for nearest neighbor estimators in the streaming setting.

\section{AUGMENTED ALGORITHM} \label{sect:augmented}

Without any structure on the sampling process it would be impossible to provide conditions on the estimator which guarantee consistency for any active learning algorithm $A$. At the same time we do not want to constrain our active learning algorithm too much. Our proposal, based on (R1) in \cite{dasgupta2012consistency}, is a simple and intuitive augmentation which is relatively inexpensive. The idea is to occasionally ignore our active learning algorithm and instead sample according to the underlying $P_X$. In query synthesis this is done directly, and in the pool setting this is done by sampling uniformly from the unlabelled data.  

\begin{algorithm}[h]
\caption{Augmented Algorithm for pool setting} \label{alg:generic}
  \SetAlgoLined
  \KwIn{Active learning algorithm $A$, number of samples $n$, probability sequence $(p_1,...,p_n)$, unlabelled data $D_n(X)$}
  \KwOut{Labelled data set $S_n$}
  $S_0 = \emptyset$ \;
  \For{$i$ from $1$ to $n$}{
    Draw an independent Bernoulli random variable $Z_i$ with $P(Z_i = 1) = p_i$\;
    \eIf{$Z_i = 1$}{Select $X_{i}$ uniformly at random from $S_{i-1}^c$}{Select $X_{i}$ according to $A(S_{i-1}, S_{i-1}^c)$}
    Query selected point and receive $Y_i$ \;
    $S_{i} = S_{i-1} \cup (X_{i}, Y_{i})$ \;
  }
\begin{remark}
In the Query Synthesis setting, if $Z_i = 1$ then our augmented algorithm will simply draw $X$ according to $P_X$ and $Y$ from $P_{Y|X}$, and the full algorithm is in the appendix.
\end{remark}
\end{algorithm}

\vspace{-0.5em}

The augmented algorithm is still an active learning algorithm. However we will refer to it as the augmented algorithm to avoid confusion with the active learning algorithm $A$ which it augments. We impose the following requirements on our sequence of $p_i$:

\vspace{-1em}

\begin{equation*}
\begin{aligned}
    p_i \searrow 0, \qquad \sum\limits_{i=1}^{\infty} p_i = \infty
\end{aligned}
\end{equation*}

\vspace{-0.5em}

The first requirement ensures that asymptotically the fraction of your data set which is sampled randomly goes to 0, and that as you collect more data, you are more likely to exploit the information you have and sample actively. The second requirement ensures we will sample at random infinitely often, even though the fraction of samples chosen randomly is asymptotically negligible. These are very similar to requirements for the $\epsilon$-greedy approach \citep{sutton1998introduction} with decaying $\epsilon_n$. 

% What properties do we want our augmented algorithm to have? One reasonable requirement is that most of our data is sampled actively. Therefore we will require that the probabilities of sampling randomly approach 0 monotonically from above.

% \begin{property}
% \label{property:goes_to_zero}
%     $p_i \searrow 0$
% \end{property}

% Property \ref{property:goes_to_zero} ensures that asymptotically the fraction of your data set which is sampled randomly goes to 0.
% %The non-increasing requirement allows this augmentation to double as a strategy to balance exploration-exploitation.
% As you collect more data, you are more likely to exploit the information you have and sample actively, as opposed to further explore your space by sampling passively. 

% Are there additional properties which are needed ensure consistency of the augmented algorithm? An obvious candidate would be that the sum of the probabilities diverges. 

% \begin{property}
% \label{property:sum_to_inf}
%     $\sum\limits_{i=1}^{\infty} p_i = \infty$
% \end{property}

% Property \ref{property:sum_to_inf} ensures we will sample at random infinitely often, even though the fraction of samples chosen randomly is asymptotically negligible. These are very similar to the $\epsilon$-greedy approach \citep{sutton1998introduction} with decaying $\epsilon_n$. 

\section{SUFFICIENCY IN THE NOISE FREE CASE} \label{sect:noiseless}

We first consider the noise free case, where we impose the following Regularity Condition on our underlying distribution: that the boundary between the two classes has $[P_X]-$measure 0: 

\begin{reg_condition}
\label{cond:separable}
    Assume we are in the noise free setting, i.e., $Y = f(X) = f^*(X)$ almost surely. Let $\mathcal{X}_0 \subset \mathcal{X}$ be $\mathcal{X}_0 = \{ x \in \mathcal{X} : \exists \: B = B_{x,r}, r>0, P_X(B) > 0, f(z) = 0 \: \forall \: z \in B \}$ and define $\mathcal{X}_1$ similarly. Then $P_X(\mathcal{X}_0 \cup \mathcal{X}_1) = 1$.
\end{reg_condition}

Under this Regularity Condition and using the augmentation in Algorithm \ref{alg:generic} classic weighted averaging estimators can all be made consistent for any base active learning algorithm $A$.

\begin{proposition}
\label{prop:noiseless_all}
    Assume Regularity Condition \ref{cond:separable}, and sample using Algorithm \ref{alg:generic} with any active learning algorithm $A$. Let $s_n = \sum\limits_{i=1}^n p_i$. Then the following estimators are consistent:
    
    \begin{itemize}
        \item The histogram estimator if $h_n \rightarrow 0, h_n^d s_n \rightarrow \infty$.
        \item $k$-nn if $\frac{k_n}{s_n}\rightarrow 0$.
    \end{itemize}
\end{proposition}

Additionally similar results can be proven for many standard bounded support kernel estimators under the condition that $h_n \rightarrow 0, h_n^d s_n \rightarrow \infty$. These conditions are almost the same as the conditions derived from Stone's Theorem under \textit{iid} sampling, except $n$ the number of samples has been replaced by $s_n$ the (expected) number of random (\textit{iid} from $P_X$) samples. 

The consistency of these is provided by a single unifying condition. The statement of the condition is somewhat technical, and we will discuss why such technicality is needed. Let $\tilde{X}_i = \tilde{X}_i(X_i, \mathbf{1}_{E_i}, V_i) = X_i \mathbf{1}_{E_i} + V_i (1 - \mathbf{1}_{E_i})$. We will define a (family of) function $g_n: \mathcal{X} \times \mathbf{R}_+ \times \mathcal{X}^n \times \{0,1\}^n \rightarrow [0,1]$ by:

\begin{equation*}
\begin{aligned}
    &g_n(x, r, \{X_i\}^n, \{\mathbf{1}_{E_i}\}^n) =\\
    & \inf\limits_{\{V_i\} \in supp(\mathcal{X})} \sum\limits_{i=1}^n W_{ni}(x, \{\tilde{X}_i\}^n) \mathbf{1}_{\tilde{X}_i \in B_{x,r}}
\end{aligned}
\end{equation*}

Note that if $\mathbf{1}_{E_i} = 0$ then the value of $X_i$ does not matter.
That is

\begin{equation*}
\begin{aligned}
    &g_n(...x_i = a ... \mathbf{1}_{E_i} = 0...) =\\ 
    &g_n(...x_i = b ... \mathbf{1}_{E_i} = 0...) \: \forall \: a,b, \{x_j\}_{j \neq i} \{\mathbf{1}_{E_j} \}_{j \neq i}
\end{aligned}
\end{equation*}

Now assume we are sampling $(Z_i, X_i)$ according to our augmented active learning algorithm, and let $E_i = \{Z_i = 1 \} \cap \{X_i \in B_{x,r} \}$. Then our Condition is the following: 

\begin{condition}
\label{property:noiseless}
    Let $X, X_i \sim P_X$ and $Z_i \sim B(p_i)$. Assume $\exists \: H_n \: s.t. \frac{H_n}{s_n} \to 0$ and  $\forall \: r > 0$:
    \begin{equation*}
    \begin{aligned}
        \underset{X}{E}\underset{Z_i}{E} \underset{X_i}{E} \big{[} g_n(X,r, \{X_i\}^n, \{\mathbf{1}_{E_i} \}^n) | \sum \mathbf{1}_{E_i} \geq H_n  \big{]} \rightarrow 1
    \end{aligned}
    \end{equation*}
\end{condition}

\begin{theorem}
\label{thrm:noiseless_case}

Assume Regularity Condition \ref{cond:separable}, that data is sampled according to Algorithm \ref{alg:generic} with any Active Learning algorithm $A$. If predictions are made with a weighted averaging estimator satisfying Condition \ref{property:noiseless} then $E \mathbf{1}_{f_{n}(X,S_n) \neq Y} \rightarrow 0$.

\end{theorem}

Condition \ref{property:noiseless} ensures that predictions are eventually made only using data within an arbitrarily small neighborhood, that those small neighborhoods are non empty, and that the weight of all data in these neighborhoods cannot be nullified by adversarial placement of additional points. The families of estimators which satisfy Stone's Theorem but not this are largely pathological and an example is given in the appendix.

\section{EXAMPLES IN THE NOISY CASE} \label{sect:noisy}

We now move beyond the noise free setting and allow for $f(x) \in [0,1]$. Following \cite{dasgupta2012consistency} we will assume a Regularity Condition on $f(x)$:

\begin{reg_condition}
\label{cond:conditional_cont}
    If the support of $P_X$ is $\{x \in \mathcal{X} : P_X(B_{x,r}) > 0 \: \forall \: r > 0\}$ then $\forall x$ in the support of $P_X$ $x$ is a continuity point of $f(x)$. 
\end{reg_condition}

This condition gives us the following property: for all $x$ except on a set of $P_X$ measure 0, and for any $\epsilon > 0$ there is a ball $B_{x,r}, P_X(B_{x,r}) > 0$ such that $|f(x) - f(z)| < \epsilon \: \forall \: z \in B_{x,r}$. We will also assume that $P_X(\{x \in \mathcal{X} : f(x) = \frac{1}{2} \}) = 0$ to remove uninteresting qualifications during statements and proofs. Under these assumptions, is Condition \ref{property:noiseless} still sufficient for consistency?
%We will show that for some estimators this is sufficient, while for others it is not. 

\subsection{Histogram Estimators}

We begin with the positive case by showing that for the histogram estimator, properties required for Condition \ref{property:noiseless} also give consistency in the noisy setting. As shown in the proof of Proposition \ref{prop:histogram}, Condition \ref{property:noiseless} hold for the histogram iff $h_n \rightarrow 0, h_n^d s_n \rightarrow \infty$, and the proof shows that if Condition \ref{property:noiseless} is satisfied, the probability of our test point falling in a partition with only $M$ data points goes to 0 for all $M < \infty$. Under our Regularity Condition \ref{cond:conditional_cont} this is sufficient for consistency

% and $h_n^d s_n \rightarrow \infty$ is required for the amount of randomly sampled data to grow at a faster rate than the number of partitions.

% \begin{proposition}
% \label{prop:hist_iff}
%     Under Regularity Condition \ref{cond:separable} the augmented algorithm with a histogram classifier is universally consistent for any base active learning algorithm $\iff h_n \rightarrow 0, h_n^d s_n \rightarrow \infty$.
% \end{proposition}

% Consistency even in the noisy case now comes from the fact that $h_n^d s_n \rightarrow \infty$ in fact gives us that probability of our test point falling in a partition with only $M$ data points goes to 0 for all $M < \infty$. And because of $h_n \rightarrow 0$ and Regularity Condition \ref{cond:conditional_cont}, eventually the majority class in every cell will be the same (in probability) as the majority class of the test point.

\begin{proposition}
\label{prop:histogram}

Under Regularity Condition \ref{cond:conditional_cont}, $h_n \rightarrow 0, h_n^d s_n \rightarrow \infty$ with a histogram classifier is consistent for any base active learning algorithm.

\end{proposition}

Therefore properties of our histogram required to satisfy Condition \ref{property:noiseless} (and therefore give consistency in the noise free case) also give consistency in the noisy case.

\subsection{Nearest Neighbor Estimators}

We now present an example where you can satisfy Condition \ref{property:noiseless} but are not consistent in the noisy setting, using nearest neighbors as our underlying estimator. In our counterexample the Bayes Risk will be $\eta$ for some $\eta > 0$ but arbitrarily small, but the risk of our augmented algorithm will be $1 - \eta$. We will present the example for $1$-NN since the intuition is strongest here, but the example generalizes when $k_n \rightarrow \infty, \frac{k_n}{s_n}\rightarrow 0$ (which is sufficient for consistency under passive sampling and when there is no noise), and we will give the corresponding theorem and guide through the proof in the appendix. Although $1$-NN is not consistent when there is noise present under passive sampling, it achieves within a factor of 2 from the optimal risk $R^*$ of the Bayes classifier \citep{cover1967nearest} whereas in our counter example it has risk close to 1.

Let $\mathcal{X} = [0,1],$ $X_i \sim U[0,1]$ and $Y_i | X_i \sim Bern(\eta), 0 < \eta < \frac{1}{2}$ (so we trivially satisfy Regularity Condition \ref{cond:conditional_cont}). Note here that the Bayes classifier $f^*(x)$ always predicts the class $0$ and has risk $\eta$. Let $f(x,S_n)$ be the prediction of a 1-NN learner at point $x$ trained on the data set $S_n$. 

This example will assume we are in the pool setting (although the translation of the example to the query synthesis setting is clear). Let $L(X):D_n(X) \rightarrow D_n(Y)$ be the look up table for the label of that data point in our pool $L(X_i) = Y_i$. We assume that acquiring unlabelled data is effectively free compared with the cost of labelling the data. In particular we will assume that $\frac{n}{m_n} \rightarrow 0$.

% We are interested in the risk of $f(x, S_n)$ when $S_n$ is selected using the augmented Algorithm \ref{alg:generic}:

% \begin{equation*}
% \begin{aligned}
%     E_{(X,Y)} E_{S_n} E_{D_n}\mathbf{1}_{f(X, S_n)\neq Y}
% \end{aligned}
% \end{equation*}

We will again use augmented Algorithm \ref{alg:generic}. However our base active learning algorithm will be a specific active learning algorithm $A^{\dagger}$ defined in the next section, which is an 'adversarial' active learning algorithm, developed purely to test the sufficiency claim of Theorem \ref{thrm:noiseless_case} when we do not assume Regularity Condition \ref{cond:separable}. We will describe informally what the algorithm does and how it achieves it's asymptotically near perfect Riskiness before presenting the proof. 

\subsubsection{Informal description of proof}

During this subsection, we will let $X_i$ be the $i^{th}$ point sampled, and let the ordered random variables $X_{(i)}$ denote ordering of the unlabelled data on the interval $[0,1]$. We will sample according to algorithm \ref{alg:generic}, with a specific active learning Algorithm $A^{\dagger}$. The active learning algorithm $A^{\dagger}$ will work in the following way: Given $S_t$ and $S_t^c$, we can define \textit{open points} as unlabelled data points who's left or right neighbor are labelled as 0:

\begin{definition}
    Let $L^t(X)$ denote the known label of point $X$ at some time $t$, with $L^t(X) = ?$ if the point is unlabelled at iteration $t$. Then a point $X_{(i)}$ is an \textit{open point} at time $t$ if $L^t(X_{(i)}) = ?, L^t(X_{(i+1)}) = 0$ or $L^t(X_{(i)}) = ?, L^t(X_{(i-1)}) = 0$.
\end{definition}

% Then at each iteration the active learning algorithm $A^{\dagger}$ (which is only defined for $X \in \mathcal{X} \subset \mathbf{R}$) samples as in Algorithm \ref{alg:1nn}:

\begin{algorithm}[h]
\caption{Adversarial Active Learning algorithm $A^{\dagger}$} \label{alg:1nn}
  \SetAlgoLined
  \KwIn{Currently labelled data $S_t$, unlabelled data $S_t^c$ }
  \KwOut{The next point to label}
  \uIf{There is at least one open point}{Sample the smallest open point.}
  \Else{Sample the unlabelled data point which is furthest from a labelled data point}
\end{algorithm}

% \begin{remark}
%     Note that the selection of the smallest open point is just a way to ensure we sample consecutive points on an interval, and any other method of ensuring that property would suffice.
% \end{remark}

Notice that whenever an open point is labelled, it is no longer an open point. If the label of that (former) open point is 0 then it (usually) creates another open point adjacent to it, and if it is 1 then it does not create a new open point. The results of this is that we will sample consecutive points in a line, creating \textit{interior points} which are labelled point who's left and right neighbor are both labelled:

\begin{definition}
    $X_{(i)}$ is an \textit{interior point} at time $t$ if $L^t(X_{(i-1)}), L^t(X_{(i)}), \text{ and } L^t(X_{(i+1)})$ are all labelled at time $t$.
\end{definition}

These interior points (plus the two points at each end) form \textit{intervals}:

\begin{definition}
    An \textit{interval} is a groups of consecutive labelled points (and we allow singleton points to be intervals of length 1).
\end{definition}

Our active learning algorithm $A^{\dagger}$ samples consecutive points until we get a point who's label is 1, which can be thought of as having 'closed off' that side of the interval. The expected distance between these interior points is $\frac{1}{m_n + 1}$. By construction all points with label $0$ are interior points, or are adjacent to open points. We will show that eventually almost all points with the label 0 are interior points. 

\begin{figure*}[ht]
	\caption{}
	\centering
	\begin{tabular}{cc}
		\includegraphics[width=0.99\textwidth,height=0.18\textheight,clip = true, trim = {20 0 20 0}]{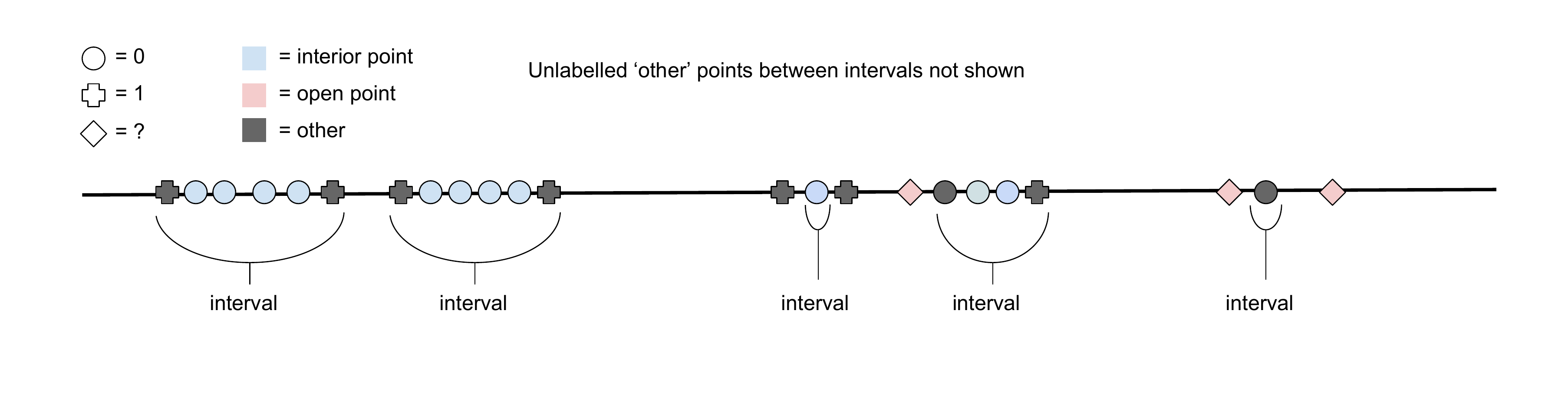}\\
	\end{tabular}
	\label{fig:1nn}
\end{figure*}

% \vspace{-5pt}

We then define the \textit{coverage} of a point as the area where they are the nearest neighbor:

\begin{definition}
    The coverage of a point $x$ is $I(x, S_n) = \int \mathbf{1}_{x = \argmin\limits_{x' \in S_n} |z-x'|} \: dz$
\end{definition}

\noindent
Note that the expected area covered by our points with label $1$ is $E[\mathbf{1}_{f(X, S_n) = 1}] = \sum\limits_{x \in S_n} E [I(x, S_n) \mathbf{1}_{L^{n}(x) = 1}]$.

% \begin{equation*}
% \begin{aligned}
%     E[\mathbf{1}_{f(X, S_n) = 1}] = \sum\limits_{x \in S_n} E [I(x, S_n) \mathbf{1}_{L^{n}(x) = 1}]
% \end{aligned}
% \end{equation*}

The coverage of all interior points is $\leq \frac{n}{m_n + 1} \rightarrow 0$. And we show that the coverage of each open point's labelled neighbor (which has label 0) also $\rightarrow 0$. Thus the area covered by points with label $1$ goes to 1, and so the risk goes to $1-\eta$ as the resulting estimator is $1 - f^*(x)$.  

\subsubsection{Formal proof}

The structure of the proof will be based around corollary \ref{cor:interior} and corollary \ref{cor:open_point}. Since all points with label 0 are either interior or adjacent to open points, we just need to control the coverage of these two types of points. First we will bound the expected coverage of $n$ interior points and see that it goes to 0. Next we will show that with high probability the number of open points will eventually be bounded. Finally we will show that each point adjacent to an open point has coverage going to 0.

% \begin{theorem}
% \label{thrm:1nn}
%     Let $X \stackrel{iid}{\sim} U(0,1)$ and let $P(Y = 1 | X) = \eta, \: 0<\eta<\frac{1}{2}$. We sample $S_n$ using augmented Algorithm \ref{alg:generic}, with properties \ref{property:goes_to_zero} and \ref{property:sum_to_inf}, and with base active learning algorithm $A^{\dagger}$ described in Algorithm $\ref{alg:1nn}$. If our estimator $f(x, S_n)$ is 1-$NN$ then $E\big{[} \mathbf{1}_{f(X, S_n) = Y} \big{]} \rightarrow 1 - \eta$.
% \end{theorem}

% We will do this by showing that the expected area covered by point with label 1 $E\big{[} \mathbf{1}_{f(X, S_n) = 1} \big{]} \rightarrow 1$, as this gives us a risk of $1-\eta$. 

Since $\frac{n}{m_n} \rightarrow 0$ the coverage of all interior points decreases faster than the number of interior points can grow. 

\begin{proposition}
\label{prop:interior}
    If $X_{(i)}$ is an interior point, then the expected area covered by that point is $E I(X_{(i)}, S_n) = \frac{1}{m_n + 1}$.
\end{proposition}

\begin{corollary}
\label{cor:interior}
    The expected area covered by all interior points approaches 0 in the limit.
\end{corollary}

Now we want to show that asymptotically the probability of there being many open points at time $n$, when we stop sampling, is small. Let $O_n$ be the number of open points at time $n$ and let $U_i$ be the change in the number of open points at time $i$ So $O_n = \sum\limits_{i=1}^n U_i$ and by construction $O_n \geq 0 \: \forall \: n$. Since the behaviour of $U_i$ is different depending on whether $O_{i-1}$ is 0 or not, we analyze $U_i$ by analyzing it's behaviour between times when it returns to 0. We will call these returns to 0 \textit{cycles}. Let $\tau_j$ be the $j^{th}$ time that $O_i = 0$, with $\tau_1 = 0$ (since with no labelled points we have no open points). We first want to show that $\tau_j < \infty \: \forall \: j$ with probability 1, that is that our number of open points returns to 0 infinitely often with probability 1. 

To do this we will bound $U_i$ by an 'idealized' process $U'_i$. This bound will only hold between cycles (since $U_i$ has different behaviour when the number of open points is 0). 

\begin{equation*}
\begin{aligned}
    % &O'_n = \sum\limits_{i=1}^n U'_i\\
    % &U'_i = W'_i \mathbf{1}_{O'_{i-1} > 0} + 2 \mathbf{1}_{O'_{i-1} = 0}\\
    &U'_i =
        \begin{cases}
    2 & \text{if } Z_i = 1 \text{ and } Y_i = 0\\ % \text{ which occurs w.p. } \frac{1}{i}(1 - \eta)\\
    -1 & \text{if } Z_i = 0 \text{ and } Y_i = 1\\ % \text{ which occurs w.p. } \eta(1 - \frac{1}{i})\\
    0 & \text{otherwise}\\
    \end{cases}
\end{aligned}
\end{equation*}

\begin{proposition}
\label{prop:driving_process}
    If $O_{i-1} \neq 0$ then $U_i \leq U'_i \: a.s.$
\end{proposition}

Note that for $i$ sufficiently large $E[U'_i] < 0$ and so $\sum\limits_{i = i_0}^{\infty} U'_i \stackrel{a.s.}{\rightarrow} -\infty$. Thus the number of open points will always return to 0 in a finite number of iterations (with probability 1).

% \begin{corollary}
% \label{cor:driving_process}
%     If $O_{i-1} \neq 0 \: \forall \:i$ then $\sum\limits_{i=1}^n U_i \leq \sum\limits_{i=1}^n U'_i \: a.s.$
% \end{corollary}

\begin{proposition}
\label{prop:hit_0_io}
    $P(O_i = 0 \: i.o.) = 1$.
\end{proposition}

So we know we return to 0 open points infinitely often with probability 1. We want to show that the probability of having a large number of open points any time during cycle $j_0$ goes to zero as $j_0 \rightarrow \infty$. 

\begin{proposition}
\label{prop:err_before_rand_one}
    Let $\tilde{T}_{1,i_0}$ be the first time after $i_0$ that $\sum\limits_{i=i_0+1}^{\tilde{T}_{1,i_0}}Z_{i} = 1$ and let $T_{1,i_0} = \tilde{T}_{1,i_0} - i_0$. Let $\tilde{T}_{2,i_0}$ be the first time after $i_0$ that $\sum\limits_{i=i_0+1}^{\tilde{T}_{2,i_0}}Y_{i} = 1$ and let $T_{2,i_0} = \tilde{T}_{2,i_0} - i_0$. Then: 
    \begin{enumerate}[i]
        \item $P(T_{1,i_0} < T_{2,i_0}) \leq p_{i_0} \frac{1}{\eta}$
        \item $P(T_{1,i_0} = T_{2,i_0}) \leq p_{i_0}$
    \end{enumerate}
\end{proposition}

The first result can be generalized to find the probability of getting $\sum Z_{i_0+t} = a$ before $\sum Y_{i_0+t} = b$. Since the $Z_i$ and $Y_i$ are all independent, these can be calculated recursively.  

\begin{corollary}
\label{cor:err_before_rand_k}
    Let $\tilde{T}^{(a)}_{1,i_0}$ be the first time after $i_0$ that $\sum\limits_{i=i_0+1}^{\tilde{T}^{(a)}_{1,i_0}}Z_{i} = a$ and let $T^{(a)}_{1,i_0} = \tilde{T}^{(a)}_{1,i_0} - i_0$. Let $\tilde{T}^{(b)}_{2,i_0}$ be the first time after $i_0$ that $\sum\limits_{i=i_0+1}^{\tilde{T}^{(b)}_{2,i_0}}Y_{i} = b$ and let $T^{(a)}_{2,i_0} = \tilde{T}^{(a)}_{2,i_0} - i_0$. If we denote $p_{i_0}^{(a,b)} = P(T^{(a)}_{1,i_0} < T^{(b)}_{2,i_0})$. Then we have the following recursive relationship:
    
    \begin{equation*}
    \begin{aligned}
        p_{i_0}^{(1,b)} &\leq p_{i_0}^{(1,1)} + (1 - p_{i_0}^{(1,1)}) p_{i_0}^{(1,b-1)} \leq b p_{i_0}^{(1,1)}\\
        p_{i_0}^{(a,1)} &\leq p_{i_0}^{(1,1)} p_{i_0}^{(a-1,1)} \leq (p_{i_0}^{(1,1)})^a\\
        p_{i_0}^{(a,b)} &= p_{i_0}^{(1,1)} p_{i_0}^{(a-1,b)} + P(T_{1,{i_0}} = T_{2,{i_0}}) p_{i_0}^{(a-1,b-1)}\\
        &+ (1 - p_{i_0}^{(1,1)} - P(T_{1,{i_0}} = T_{2,{i_0}})) p_{i_0}^{(a,b-1)}\\
        & \leq p_{i_0}^{(1,1)} p_{i_0}^{(a-1,b)} + \eta p_{i_0}^{(1,1)} p_{i_0}^{(a-1,b-1)} + p_{i_0}^{(a,b-1)}
    \end{aligned}
    \end{equation*}
    
    In particular we have that $p_{i_0}^{(a,b)} \leq 3^{a+b} (p_{i_0}^{(1,1)})^a$.
\end{corollary}

This shows that the probability of increasing beyond 4 open points before dropping back down to 0 open points $p_{i_0}^{(2,4)}$ is decreasing to 0.

\begin{lemma}
\label{lemma:few_open}
    $P(O_n > 4) \rightarrow 0$ as $n \rightarrow \infty$.
\end{lemma}

We already know that points with label 0 which are not adjacent to open points are interior points. So we just need to show the contribution from the (up to 4) non-interior points with label 0 is shrinking to 0. We will do this by showing that the maximal distance between two intervals goes to 0. 

\begin{proposition}
\label{prop:open_point}
    Let $d_t$ be the maximum of all distances between consecutive intervals at time $t$. Then $d_n \stackrel{a.s.}{\rightarrow} 0$.
\end{proposition}

\begin{corollary}
\label{cor:open_point}
    The coverage of labelled points adjacent to open points $\stackrel{a.s.}{\rightarrow 0}$.
\end{corollary}

With corollaries \ref{cor:interior} and \ref{cor:open_point} we can now prove Theorem \ref{thrm:1nn}.

\begin{theorem}
\label{thrm:1nn}
    Let $X \stackrel{iid}{\sim} U(0,1)$ and let $f(x) = \eta, \: 0<\eta<\frac{1}{2}$. We sample $S_n$ using augmented Algorithm \ref{alg:generic}, and with base active learning algorithm $A^{\dagger}$ described in Algorithm $\ref{alg:1nn}$. If our estimator $f(x, S_n)$ is 1-$NN$ then $E\big{[} \mathbf{1}_{f(X, S_n) = Y} \big{]} \rightarrow 1 - \eta$.
\end{theorem}

As stated earlier, this counterexample persists even if you require $k_n \rightarrow \infty$ and only stipulate that $\frac{k_n}{s_n}\rightarrow 0$, which is required by Condition \ref{property:noiseless} (and which gives consistency if our data is sampled passively). Although the results is infinitesimally weaker, and the definitions and techniques are more complex, the main idea behind the proof is the same, and the proof can be found in the appendix.

\begin{theorem}
\label{thrm:knn}
    Let $X \stackrel{iid}{\sim} U(0,1)$ and let $f(x) = \eta, \: 0<\eta<\frac{1}{2}$ and fix $\epsilon > 0$. We create our labelled training set $S_n$ using augmented Algorithm \ref{alg:generic}, with $P(Z_i = 1) = \frac{1}{i}$, and with base active learning algorithm $A^{\dagger}$ described in Algorithm $\ref{alg:1nn}$. If our estimator is $k$-NN then $\exists \: \{k_n\}_{n=1}^{\infty}$ which satisfies Condition \ref{property:noiseless} and $\liminf E\big{[} \mathbf{1}_{f_{n}(x, S_n) = Y} \big{]} \geq 1-\eta - \epsilon$.
\end{theorem}

\section{SUFFICIENCY FOR BOUNDED SUPPORT ESTIMATORS} \label{sect:bounded}

We now aim to extract the properties of the histogram estimator which make it immune to the type of attack used in the nearest neighbor counterexample. Our conditions will assume that the weight functions $W_{ni}(x, S_n(X))$ take a simplified form, where which training points have non-zero weight only depends on $x, X_i$ and $n$. Similar to Condition \ref{property:noiseless}, these conditions will be complex to state mathematically, but will have interpretable effects. 

\begin{condition}
\label{property:bounded_sup}
\begin{equation*}
\begin{aligned}
    &1) \: W_{ni}(x, S_n(X)) = \frac{w_n(x, X_i)}{\sum\limits_j w_n(x, X_j)}\\ &2) \: \text{if } supp_n(x) = \{y \in \mathcal{X} : w_n(x, y) > 0 \} \text{ then } \\ 
    &diam(supp_n(x)) \rightarrow 0\\ 
    &3) \: w_n(x, y) \leq K \: \forall \: n, x, y\\ 
    &4) \: \sum\limits_{i=1}^n w_n(X, X_i)Z_i \CinP \infty %\quad (X, X_i \sim P_X).
\end{aligned}
\end{equation*}
\end{condition}

By enforcing this structure on $W_{ni}(x)$, we allow the unnormalized weight of each point to depend only on the location of the training point $X_i$ and the test point $x$, preventing the relative weight of a point from being affected after the label has been observed. By forcing the support to shrink in size we ensure that the method is sufficiently local. Finally by bounding the maximum relative weight of any single point and requiring that the relative weights of our randomly sampled points is unbounded (in probability), we ensure that no finite amount of actively sampled data can overwhelm our passively sampled data. Note that this implicitly requires that $\sum\limits_{i=1}^n \mathbf{1}_{X_i \in supp_n(X), Z_i = 1} \CinP \infty$, which is the key property in the proof of Proposition \ref{prop:histogram}. Although this generalization only includes certain partition estimators and bounded support regular kernel estimators who's kernel function is also bounded away from 0 on their support, it allows for a proof of consistency in the noisy case which is illuminating.

\begin{theorem}
\label{thrm:bounded_sup_consistency}

Assume our classifier and augmented algorithm satisfy Condition \ref{property:bounded_sup}. Then under Regularity Condition \ref{cond:conditional_cont} our estimator is consistent for any active learning algorithm $A$.

\end{theorem}

\section{CONCLUSION AND FURTHER DIRECTIONS} \label{sect:conclusion}

We have seen that in the noiseless setting under mild conditions classical weighted averaging estimators are consistent with a small amount of data sampled randomly. However once even a little noise is introduced there is a bifurcation, where some estimators such as the histogram retain this consistency while others such as $k$-nn can be made highly inconsistent even if they are consistent in the noiseless case. The structure of the counterexample in Section \ref{sect:noisy} and the Condition in Section \ref{sect:bounded} suggests this divergence stems from how dramatically the relative weight of a data point can be affected after it's label has been observed, and how few data points determine the final prediction. This explains why both adversarial sampling and label noise were needed to highlight the differences in behaviour. As seen in the $1$-NN counterexample (the structure of which can also give counterexamples for unbounded kernel estimators with sufficiently quickly shrinking $h_n$), if the influence of one data point can be too easily manipulated (after observing it's label) by the placement of other data points, we can get inconsistency even with our randomly sampled data. Condition \ref{property:bounded_sup} strongly protect against this, and less strenuous conditions can likely be found for local averaging estimators. However more interestingly the intuition behind these properties may provide guidance when using more modern estimators, and exploring and formalizing this is the subject of future work.

One direction would be to explore whether this disjunction in the vulnerability of different estimators is mirrored for more advanced methods. Under passive sampling SVMs and Random Forests are both competitive classifiers \citep{caruana2006empirical}, but given the similarities between SVM and Nearest Neighbors, and Random Forests and Histograms, their guarantees may be very different under active sampling. Another potential avenue would be finding ways to adapt complex methods to maintain consistency under Algorithm \ref{alg:generic} or similar schemes. For example the soft-margin SVM dual form optimization variables $\alpha_i$ encode the influence of a data point on the prediction of nearby points. The high level ideas in Condition \ref{property:bounded_sup} suggest additional constraints (such as $\max_{i} \alpha_i - B_n \sum \alpha_i \leq 0, B_n \rightarrow 0$) may result in a version of the SVM which is more robust under a similar augmented active learning algorithm. Finally it would be interesting to see how these Conditions change if we put constraints on the underlying active learning algorithm being augmented. 

% We have shown that a simple possible fix to consistency only works when there is no noise, and that when noise is present we can still produce an estimator who's prediction is the opposite of the Bayes estimator $f^*(x)$. In order for the fix to work when there is noise present we required the active learning algorithm to use separate data for the data selection part and for the final predictor. 

% Since active learning is useful exactly when labelled data is precious, requiring some labelled data be excluded from use by the final predictor makes this fix impractical. Therefore finding more palatable ways of ensuring consistency, either through augmenting existing algorithms or putting conditions on those algorithms, is still a very interesting and challenging open problem in active learning. Additionally these results are only for $k_n$-NN methods, but it would be useful to know if the problem is similarly difficult for other consistent methods. 

\subsubsection*{Acknowledgements}
JG acknowledges the support of NSF via grant DMS-1646108. AT acknowledges the support of a Sloan Research Fellowship.

\bibliographystyle{apa}
\bibliography{refs}

\clearpage

\section{Appendix A: COUNTEREXAMPLE FOR $k_n \rightarrow \infty$}

In order to more accurately mirror the consistency conditions under passive sampling we now add the requirement that $k_n \rightarrow \infty$.

\begin{property}
\label{property:kn_inf}
    $k_n \rightarrow \infty, \frac{k_n}{s_n}\rightarrow 0$
\end{property}

Our counterexample will be similar to in the $1$-NN case, but we will work with $p_i = \frac{1}{i}$ instead of a generic $p_i$. The only difference will be in the definition of an open point, which will need to be generalized to depend on $k_n$. If $k_n = k$ then an open point will be an unlabelled point with at least one labelled neighbour, and without $\lfloor \frac{k}{2} \rfloor + 1 = k'$ 1 labels in a row to the left or right. 

\begin{definition}
    A point $X_{(i)}$ is an \textit{open point} at time $t$ if $L^t(X_{(i)}) = ?, L^t(X_{(i+1)}) \in \{0,1\}$ and $(L^t(X_{(i+1)},...,L^t(X_{(i+k')}) \neq \mathbf{1}_{k'}$ or $L^t(X_{(i)}) = ?, L^t(X_{(i-1)}) \in \{0,1\}$ and $(L^t(X_{(i-1)},...,L^t(X_{(i-k')}) \neq \mathbf{1}_{k'}$, where $\mathbf{1}_{k'}$ is a $k'$-vector of 1's.
\end{definition}

Note that when $k = 1$ this is the same as our previous definition, and that intervals will have the same effect as before, where two consecutive intervals without any open points between them will cause any test points between them to be predicted 1. Similarly we will extend our definition of coverage to be the area where a set of size $k$ are the $k$ closest labelled points.

\begin{definition}
    Let $C_{k}(x, S_n) = \argmin\limits_{C \subset S_n, |C| = k} \sum\limits_{x' \in C} |x' - x|$. Then the $k$\textit{-coverage} of a set $C$ is $I_{k}(C, S_n) = \int \mathbf{1}_{C = C_{k}(z, S_n)}  \: dz$
\end{definition}

Note that the only sets with non-zero coverage are sets of consecutive (within $S_n$) labelled points. This again partitions the real line and we get a decomposition of our expected coverage by 1.

\begin{equation*}
\begin{aligned}
    &E[\mathbf{1}_{f_{n}(X, S_n)=1}] = \sum\limits_{C \in \mathcal{C}_k} E \Big{[} I_{k}(C, S_n) \mathbf{1}_{\sum\limits_{x \in C} L^{n}(x) \geq k'} \Big{]}\\
    &(\text{where } 1 + ? = 1, 0 + ? = 0)\\
    &\mathcal{C}_k = \{C \subset S_n : C = \{X_{(i)},...,X_{(i+k)}\} \}\\
    &\text{Where this ordering is only over } X \in S_n\\
\end{aligned}
\end{equation*}

\begin{figure*}[h!]
	\caption{}
	\centering
	\begin{tabular}{cc}
		\includegraphics[width=0.99\textwidth,height=0.2\textheight,clip = true, trim = {20 0 20 0}]{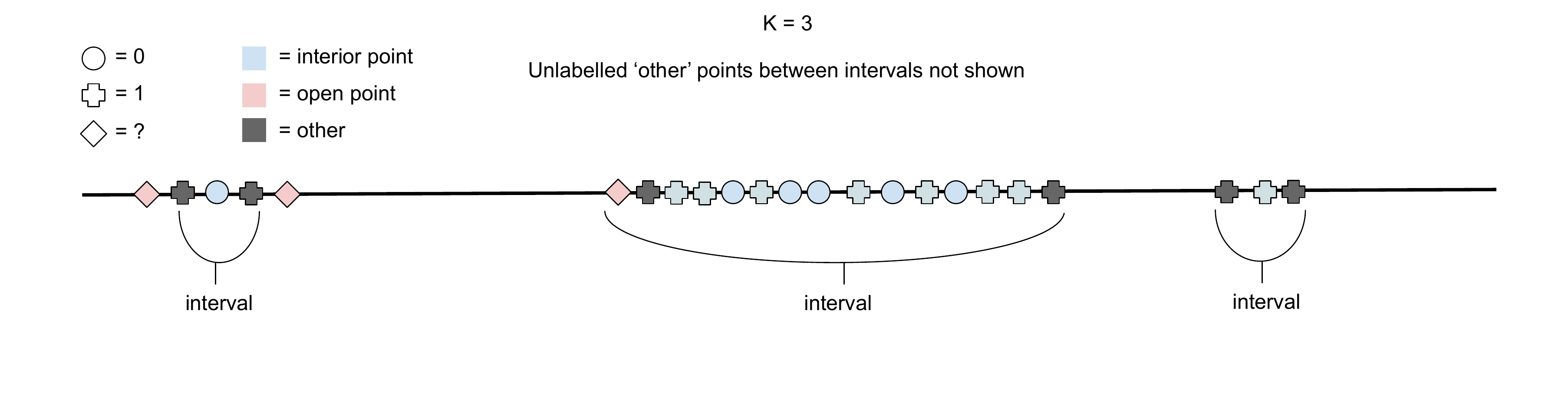}\\
	\end{tabular}
	\label{fig:knn}
\end{figure*}

For each $k$ fixed the proof will follow largely the same structure; although getting $k'$ 1's in a row is a much lower probability event than just getting a single 1, the probability is still constant (for fixed $k$), where as the probability of sampling randomly is shrinking, and so eventually the number of open points will be small. 

% \begin{theorem}
% \label{thrm:knn}
%     Let $X \stackrel{iid}{\sim} U(0,1)$ and let $P(Y = 1 | X) = \eta, \: 0<\eta<\frac{1}{2}$ and fix $\epsilon > 0$. We create our labelled training set $S_n$ using augmented Algorithm \ref{alg:generic}, with $P(Z_i = 1) = \frac{1}{i}$, and Property \ref{property:kn_inf}, and with base active learning algorithm $A^{\dagger}$ described in algorithm $\ref{alg:1nn}$. If our estimator is $f_{n}(x, S_n)$ then $\exists \: \{k_n\}_{n=1}^{\infty}$ which satisfies the above properties and $\liminf E\big{[} \mathbf{1}_{f_{n}(x, S_n) = Y} \big{]} \geq 1-\eta - \epsilon$.
% \end{theorem}

Our strategy will be very similar to in the $1$-NN case, which was to show that the expected area covered by point with label 1 $E\big{[} \mathbf{1}_{f_{n}(X, S_n) = 1} \big{]} \rightarrow 1$, as this gives us a risk of $1-\eta$. 

We again use $U_i$ to denote the change in the number of open points, and will again use an idealized version $U'_i$ which dominates $U_i$ to simplify analysis.

% \vspace{-10pt}
% \begin{enumerate}
%     \item After drawing a $Z_i = 1$ the counts for all open points are reset. So if you have two 1's in a row and then draw randomly, we do not 'remember' the two 1's whenever you go back to working on that open point.
%     \item Between consecutive runs of $k'$ 1's you must draw at least a single 0.
% \end{enumerate}
% \vspace{-10pt}

% We again do this to remove complicated and low impact long range dependency. As with before by making these simplifying assumptions we are only removing events which would reduce the number of open points.

\begin{equation*}
\begin{aligned}
    % &O'_n = \sum\limits_{i=1}^n U'_i\\
    % &U'_i = W'_i \mathbf{1}_{O'_{i-1} > 0} + 2 \mathbf{1}_{O'_{i-1} = 0}\\
    &U'_i =
        \begin{cases}
    2 & \text{if } Z_i = 1 \text{ and } Y_i = 0\\ % \text{ which occurs w.p. } \frac{1}{i}(1 - \eta)\\
    -1 & \text{if } Z_j = 0 \: \forall \: j \in \{i,...,i-k' \} \text{ and }\\ 
    &Y_j = 1 \: \forall \: j \in \{i,...,i-k' \} \text{ and } \\ % \text{ which occurs w.p. } \eta(1 - \frac{1}{i})\\
    &Y_{i-k'-1} = 0\\
    0 & \text{otherwise}\\
    \end{cases}
\end{aligned}
\end{equation*}

The following propositions all have the same proofs as in the $1$-NN case, since $P(U'_i = 2) \rightarrow 0$ and $P(U'_i = -1) \rightarrow \eta^{k'}(1-\eta)$. Our $U'_i$ are no longer independent, but they do have finite range independence, and so we still have a SLLN for them. 

\begin{proposition}
\label{prop:driving_process_kn}
    If $O_{l} \neq 0$ for $l \in \{i-k'-2,...,i-1\}$ then $U_i \leq U'_i \: a.s.$
\end{proposition}

\begin{proposition}
\label{prop:hit_0_io_kn}
    $P(O_i = 0 \: i.o.) = 1$.
\end{proposition}

Now we will get the equivalent to proposition \ref{prop:err_before_rand_one}.

\begin{proposition}
\label{prop:close_before_rand_k}

    Assume $P(Z_i = 1) = \frac{1}{i}$, $P(Y_i = 1) = \eta$ and $k$ fixed. Let $\tilde{T}_{1,i_0}$ be the first time after $i_0$ that $\sum\limits_{i=i_0+1}^{\tilde{T}_{1,i_0}}Z_{i} = 1$ and let $T_{1,i_0} = \tilde{T}_{1,i_0} - i_0$. Let $\tilde{T}_{2,i_0}$ be the first time after $i_0$ that $\sum\limits_{i=i_0 + \tilde{T}_{2,i_0} - k'}^{\tilde{T}_{2,i_0}}Y_{i} = k'$ and let $T_{2,i_0} = \tilde{T}_{2,i_0} - i_0$. Then: 
    \begin{enumerate}[i]
        \item $P(T_{1,i_0} < T_{2,i_0}) \leq \frac{c}{2 \sqrt{i_0}} \big{[} \frac{1}{1-\eta} \big{(} \frac{1}{\eta^{k'}} - 1 \big{)} \big{]}$
        \item $P(T_{1,i_0} = T_{2,i_0}) \leq \frac{1}{i_0}$
    \end{enumerate}

\end{proposition}

And if we generalize we have the same recursive relationships.

\begin{corollary}
\label{cor:close_before_rand_k}
    Let $\tilde{T}^{(a)}_{1,{i_0}}$ be the first time after ${i_0}$ that $\sum\limits_{i={i_0}+1}^{\tilde{T}^{(a)}_{1,{i_0}}}Z_{i} = a$ and let $T^{(a)}_{1,{i_0}} = \tilde{T}^{(a)}_{1,{i_0}} - {i_0}$. Let $\tilde{T}^{(b)}_{2,{i_0}}$ be the first time after ${i_0}$ that we've had $k'$ out of the last $k$ queries be error terms on $b$ disjoint occasions (so starting over each time) and let $T^{(a)}_{2,{i_0}} = \tilde{T}^{(a)}_{2,{i_0}} - {i_0}$. If we denote $p_{i_0}^{(a,b)} = P(T^{(a)}_{1,{i_0}} < T^{(b)}_{2,{i_0}})$. Then we have the following recursive relationship:
    
    \begin{equation*}
    \begin{aligned}
        &p_{i_0}^{(1,b)} \leq p_{i_0}^{(1,1)} + (1 - p_{i_0}^{(1,1)}) p_{i_0}^{(1,b-1)} \leq b p_{i_0}^{(1,1)}\\
        &p_{i_0}^{(a,1)} \leq p_{i_0}^{(1,1)} p_{i_0}^{(a-1,1)} \leq (p_{i_0}^{(1,1)})^a\\
        &p_{i_0}^{(a,b)} \leq p_{i_0}^{(1,1)} p_{i_0}^{(a-1,b)} + \frac{1}{{i_0}} p_{i_0}^{(a-1,b-1)} + p_{i_0}^{(a,b-1)}\\
    \end{aligned}
    \end{equation*}
    
    In particular we have that $p_{i_0}^{(a,b)} \leq 3^{a+b} (p_{i_0}^{(1,1)})^a$
\end{corollary}

Now we have a (slightly stronger) equivalent to lemma \ref{lemma:few_open}.

\begin{lemma}
\label{lemma:under_6_as}

For any $k$, $P(O_n > 6 \: i.o.) = 0$.

\end{lemma}

This means $\mathbf{1}_{O_n > 6} \stackrel{a.s.}{\rightarrow} 0$. Therefore by an equivalent definition of almost sure convergence \citep{chung2001course} $\exists \: n_{k,\epsilon} \: s.t. \: P(\mathbf{1}_{O_n > 6} \neq 0 \: \forall n \geq n_{k,\epsilon}) \leq \epsilon$. Of course we have no way of knowing what $n_{k, \epsilon}$ is for each values of $k,\epsilon$, but we know they exist. Therefore we will allow $k_n$ to increase in the following manner (which we denote $k(n,\epsilon)$): 

\begin{itemize}
    \item $k_n = 1$ for $n < n_{2,\epsilon}$
    \item $k_n = 2$ for $n \in [n_{2,\epsilon}, n_{3,\epsilon}]$
    \item ...
    \item $k_n = k$ for $n \in [n_{k,\epsilon}, n_{k+1,\epsilon}]$
\end{itemize}

Of course we also need to satisfy $\frac{k_n}{s_n} \rightarrow 0$ and so we can just take $k_n = \min(k(n, \epsilon), \log\log(n))$.

The rest of the proof follows as in the $1$-NN case.

\begin{proof}[Proof of theorem \ref{thrm:knn}]
    Let the sequence $\{k_n\}_{n=1}^{\infty}$ be as described above. Then for $n \geq n_2$ we know that when we have finished taking our $n$ samples, $P(\mathbf{1}_{O_n > 6} \neq 0 \: \forall n \geq n_{k,\epsilon}) \leq \epsilon$.

    We therefore split our expected coverage with 1 into $E[\mathbf{1}_{f_{n}(X, S_n) = 1}] = E[\mathbf{1}_{f_{n}(X, S_n) = 1} \mathbf{1}_{O_n > 6}] + E[\mathbf{1}_{f_{n}(X, S_n) = 1} \mathbf{1}_{O_n \leq 6}]$. Trivially $E[\mathbf{1}_{f_{n}(X, S_n) = 1} \mathbf{1}_{O_n > 6}] \geq 0$. So we focus on $E[\mathbf{1}_{f_{n}(X, S_n) = 1} \mathbf{1}_{O_n \leq 6}]$.
    
    %= P(O_n \leq 6) E[\mathbf{1}_{f_{n}(X, S_n) \neq Y} | O_n \leq 6].
    
    Again all area which may be predicted as 0 are on the interior of intervals, or are covered by the points next to open points. The expected $k$-coverage of a set $C \in \mathcal{C}_{k_n}$ of all interior points is again $\frac{1}{m_n + 1}$ and there are fewer than $n$ such sets. This leaves up to 6 $C$ that are not all interior points, and which could have $\sum\limits_{x \in C}L^n(x) < k'$. These are all on the edges of intervals, and the lengths between intervals are still approaching 0 with probability 1 by proposition \ref{prop:open_point}. Thus we have $E[\mathbf{1}_{f_{n}(X, S_n)=1} \mathbf{1}_{O_n \leq 6}] \rightarrow 1-\epsilon$, and so $E[\mathbf{1}_{f_{n}(X, S_n)=1}] \geq 1-\epsilon$, which gives us $E[\mathbf{1}_{f_{n}(X, S_n) \neq Y}] \geq (1-2\eta)(1 - \epsilon) + \eta \geq 1-\eta - \epsilon$.
\end{proof}

\section{Appendix B: ADDITIONAL EXAMPLES AND PROOFS}

\subsection{Counterexample for Condition \ref{property:noiseless}}

The goal of Condition \ref{property:noiseless} is to ensure we get consistency just from the small amount of randomly sampled data, and to exclude estimators which can be 'tricked' by reducing the weight of the randomly sampled data in an adversarial manner. One example would be a version of the histogram estimator where data points which are within a certain (decreasing) distance of another data point are given $W_{ni}(x) = 0 \: \forall \: x$. If the radius decreases quickly enough then under random sampling the fraction of data which is nullified will be vanishing and so this estimator would behave the same way as the standard histogram. However an adversarial active learning algorithm can give all the randomly sampled data weight of 0 and so the augmentation has effectively no effect. 

\subsection{Augmented Algorithm for query synthesis}

\begin{algorithm}[h!]
\caption{Augmented Algorithm for query synthesis} \label{alg:generic_query_synth}
  \SetAlgoLined
  \KwIn{Active learning algorithm $A$, number of samples $n$, probability sequence $(p_1,...,p_n)$, underlying marginal distribution $P_X$}
  \KwOut{Labelled data set $S_n$}
  $S_0 = \emptyset$ \;
  \For{$i$ from $1$ to $n$}{
    Draw an independent Bernoulli random variable $Z_i$ with $P(Z_i = 1) = p_i$\;
    \eIf{$Z_i = 1$}{Draw $X_{i}$ from $P_X$}{Select $X_{i}$ according to $A(S_{i-1})$}
    Query selected point and receive $Y_i$ \;
    $S_{i} = S_{i-1} \cup (X_{i}, Y_{i})$ \;
  }
\end{algorithm}

\subsection{Sufficiency in the noise free case}

Why is Condition \ref{property:noiseless} our requirement? Fix $X = x$ and let $\phi$ be the distribution on $(X_1,...,X_n)$ induced by our augmented AL algorithm. By the definition of $E_i$ we have that $\mathbf{1}_{E_i} = 1 \implies Z_i = 1$ and so:

\begin{equation*}
\begin{aligned}
    &\underset{Z_i \sim B(p_i)}{E} \Big{(} \underset{X \sim \phi}{E} \big{[} g_n(x,r, \{X_i\}^n, \{\mathbf{1}_{E_i} \}^n) | \{Z_i\}^n \big{]} \Big{)}\\ &= \underset{Z_i \sim B(p_i)}{E} \underset{X_i \sim \mu}{E} \big{[} g_n(x,r, \{X_i\}^n, \{\mathbf{1}_{E_i} \}^n) \big{]}
\end{aligned}
\end{equation*}

And from this and the definition of $g_n$ we have that $\forall \: k$:

\begin{equation*}
\begin{aligned}
    &\underset{Z_i \sim B(p_i)}{E} \underset{X \sim \phi}{E} \big{[} \sum\limits_{i=1}^n W_{ni}(x, \{{X}_i\}^n) \mathbf{1}_{{X}_i \in B_{x,r}} | \sum \mathbf{1}_{E_i} = k \big{]}\\
    &\geq \underset{Z_i \sim B(p_i)}{E} \underset{X_i \sim \mu}{E} \big{[} g_n(x,r, \{X_i\}^n, \{\mathbf{1}_{E_i} \}^n) | \sum \mathbf{1}_{E_i} = k \big{]}
\end{aligned}
\end{equation*}

\begin{proof}[Proof of Theorem \ref{thrm:noiseless_case}]
    
    We want to show that $\int E_{S_n} \big{[} (f_n(x,S_n) - f(x))^2 \big{]} P_X(dx) \rightarrow 0$. 
    
    \begin{equation*}
    \begin{aligned}
        &\int E_{S_n} \big{[} (f_n(x,S_n) - f(x))^2 \big{]} P_X(dx)\\ 
        &= \int E_{S_n} \big{[} (\sum W_{ni}(x) f(x_i) - f(x))^2 \big{]} P_X(dx)\\
        &\leq 2 \int E_{S_n} \big{[} (\sum W_{ni}(x) f(x_i) - \sum W_{ni}(x)f(x))^2 \big{]} P_X(dx)\\ 
        &+ 2 \int E_{S_n} \big{[} (f(x)[\sum W_{ni}(x) - 1])^2 \big{]} P_X(dx)
    \end{aligned}
    \end{equation*}
    
    We will work on bounding the first term since the second term trivially goes to 0 due to Condition \ref{property:noiseless}.
    
    \begin{equation*}
    \begin{aligned}
        &\int E_{S_n} \big{[} (\sum W_{ni}(x) f(x_i) - \sum W_{ni}(x)f(x))^2 \big{]} P_X(dx)\\ 
        &\leq \int E_{S_n} \big{[} \sum W_{ni}(x) (f(x_i) - f(x))^2 \big{]} P_X(dx)\\
        &\text{Define } \mathcal{X}^{(\delta)} = \{x : |f(B_{x,\delta})| = 1 \},\\ 
        &\delta_{\epsilon} = \sup \delta \: s.t. \: P_X(\mathcal{X}^{(\delta)}) \geq 1- \epsilon.\\
        % &\text{This set is always measureable since it is the union of open balls centered at points in $(\mathcal{X}_0 \cup \mathcal{X}_1)^c$}.\\
        &\leq \int_{\mathcal{X}^{(\delta_{\epsilon})}} E_{S_n} \big{[} \sum W_{ni}(x) (f(x_i) - f(x))^2 \big{]} P_X(dx) + \epsilon\\
        &\text{Let } S_n = S_n^{(a)} \cup S_n^{(r)} \text{ where the first is actively selected}\\ 
        &\text{data and the second is the randomly selected.}\\
        &|B_{x, \delta_{\epsilon}} \cap S_n^{(r)}| \rightarrow P_X(B_{x,\delta_{\epsilon}}) \sum\limits^n p_i.\\ 
        &\text{For each } x \: \exists \: n_0(x) \: s.t. \: P_X(B_{x,\delta_{\epsilon}}) \sum\limits^n p_i \geq H_n \: \forall n \geq n_0(x).\\
        &\exists n_0 \: s.t. \: P_X(\{ x : P_X(B_{x,\delta_{\epsilon}}) \sum\limits^n p_i \leq H_n \}) \leq \epsilon.\\
    \end{aligned}
    \end{equation*}
    
    Let $n$ be sufficiently large and denote the intersection of the complement of the above set with $\mathcal{X}^{(\delta_{\epsilon})}$ by $\tilde{\mathcal{X}}$.
    
    \begin{equation*}
    \begin{aligned}
        &\leq \int_{\tilde{\mathcal{X}}} E_{S_n} \big{[} \sum W_{ni}(x) (f(x_i) - f(x))^2 \big{]} P_X(dx) + 2\epsilon.\\ 
            &\text{ Let $F_n$ be the event that }|B_{x, \delta_{\epsilon}} \cap S_n^{(r)}| \geq H_n.\\
            &= \int_{\tilde{\mathcal{X}}} P(F_n) E_{S_n} \big{[} \sum W_{ni}(x) (f(x_i) - f(x))^2 | F_n \big{]}\\
            &+ P(F_n^c) E_{S_n} \big{[} \sum W_{ni}(x) (f(x_i) - f(x))^2 | F_n^c \big{]} P_X(dx) + 2\epsilon\\
            &\leq \int_{\tilde{\mathcal{X}}} E_{S_n} \big{[} \sum W_{ni}(x) (f(x_i) - f(x))^2 | F_n \big{]} P_X(dx) + 3 \epsilon\\
            &\text{For }n \geq n_1 \text{ since } P_X(B_{x, \delta_{\epsilon}}) \text{ bounded away from 0}).\\
            &\leq \int_{\tilde{\mathcal{X}}} E_{S_n^{(r)}} \big{[} \sup\limits_{S_n^{(a)}} \sum W_{ni}(x) \mathbf{1}_{||X_i - x|| \geq \delta_{\epsilon}} | F_n \big{]} P_X(dx) + 3 \epsilon\\
            &\rightarrow 3 \epsilon.
    \end{aligned}
    \end{equation*}
\end{proof}

\begin{proof}[Proof of Proposition \ref{prop:noiseless_all}]

In the proof of part $i)$ we will actually prove that the condition is if-and-only-if since this will be needed in section 4. 
    
Let $N_n^{(R)} = \sum Z_i$ be the number of labelled points selected randomly. Let $A_n(x)$ denote the cell containing the point $x$ and let $N_n(x) = \sum \mathbf{1}_{X_i \in A_n(x)}$ be the number of labelled points in the same cell as $x$, and let $N_n^{(R)}(x) = \sum \mathbf{1}_{X_i \in A_n(x)}Z_i$ be the number of labelled points in the same cell as $x$ which were selected randomly.

We first prove the forward direction by showing we satisfy Condition \ref{property:noiseless}. Let $H_n = \Big{\lfloor} \frac{\sqrt{s_n}}{\sqrt{h_n^d}} \Big{\rfloor}$, noting that $\frac{H_n}{s_n} = \frac{1}{\sqrt{h_n^d s_n}} \rightarrow 0$ and $H_n h_n^d = \sqrt{h_n s_n} \rightarrow \infty$. Since $h_n \rightarrow 0$, for any $r>0$ eventually the entire cell a data point is in will be within $r$ of the point. Then repeat the proof of Theorem 6.2 in \cite{devroye2013probabilistic}, replacing $n$ with $H_n$, to show that $P(N_n^{(R)}(X) \leq M) \rightarrow 0 \: \forall \: M < \infty$. This completes the proof since a non-empty histogram has $\sum W_{ni}(x) = 1$, and for $n$ sufficiently large all the training points with non-zero weight will be within $r$.

If $h_n \not\to 0$ then clearly the Condition cannot hold for $r$ sufficiently small as the ball $B_{x,r}$ can be made arbitrarily small compared to the minimum size of the cell.

If $h_n^d s_n \rightarrow 0$, then the number of cells is growing at a faster rate than the number of randomly sampled data points, and if our active algorithm just samples the nearest neighbor to the point last sampled, then the majority of cells would end up with no data and would thus have $\sum W_{ni}(x) = 0$. 

This leaves us with the case where $h_n^d s_n \in [\alpha_1, \alpha_2], \: 0 < \alpha_1 \leq \alpha_2 < \infty$. We can study this using the theory of Random Allocations \cite{kolchin1978random}, which characterizes the properties of counts of urns with $k$ balls after $n$ balls are placed iid into urns. If we have a uniform distribution on $\mathcal{X}$ then we are in the Central Domain with equiprobable allocation, and from Theorem 1 (p.18) of \cite{kolchin1978random}, we have that for any $\epsilon > 0$, for $n$ sufficiently large $P(N_n^{(R)}(X) = 0) \geq e^{-h_n^d (1+\epsilon)s_n}$ almost surely. This is because the number of cells with no randomly sampled points is normally distributed around $\frac{1}{h_n} e^{-h_n^d (1+\epsilon)s_n}$ with variance that is $O(\frac{1}{h_n})$. Thus as above satisfying the Condition is impossible if, for example, our active algorithm just samples the nearest neighbor to the point last sampled. 

For part $ii)$ Condition \ref{property:noiseless} is satisfied with $H_n = k_n$ as long as for any fixed $r > 0$ and any $x$, random sampling puts more than $k_n$ data points $B_{x, r}$, and this is proved in Lemma 1 in \citet{dasgupta2012consistency}. 

\end{proof}

\subsection{EXAMPLES IN THE NOISY CASE}

\begin{proof}[Proof of Proposition \ref{prop:histogram}]

By Regularity Condition \ref{cond:conditional_cont}, for $n$ large enough all but an $\epsilon > 0$ $P_X$-measure of cells will be such that $f^*(x_1) = f^*(x_2) \: \forall \: x_1, x_2 \in A_{nj}$, where $A_{nj}$ is an arbitrary cell in our histogram. Therefore we need to show that $P(N_n^{(R)}(X) \leq M) \rightarrow 0 \: \forall \: M$. But if we fix $N_n^{(R)}$, this is exactly the result in Theorem 6.2 of \cite{devroye2013probabilistic}, with $n$ replaced by $N_n^{(R)}$. And by Levy's extension to Borel-Cantelli \cite{williams1991probability} we know that for any $\delta > 0$, for $n$ sufficiently large $N_n^{(R)} \in [(1-\delta)s_n, (1+\delta)s_n]$ with probability 1. Thus with probability 1 we have that $h_n^d N_n^{(R)} \rightarrow \infty$, so the conditions of Theorem 6.2 in \cite{devroye2013probabilistic} hold with probability 1, ensuring that $P(N_n^{(R)}(X) \leq M) \rightarrow 0 \: \forall \: M$ thereby completing the proof.
    
\end{proof}

\subsection{NEAREST NEIGHBOR COUNTEREXAMPLE}

\begin{proof}[Proof of proposition \ref{prop:interior}]

Since $X_{(i)}$ is an interior point, both of these neighbors are labelled, and so $X_{(i)}$ will only be the closest point on an area half of the distance between its neighbors on either side. Since the $X_i \sim U(0,1)$ the expected distance between $X_{(i)}$ and its neighbor on either side is $\frac{1}{m_n + 1}$. Therefore the expected coverage is $\frac{1}{m_n + 1}$.
\end{proof}

\begin{proof}[Proof of corollary \ref{cor:interior}]
    Each interior point covers $\frac{1}{m_n + 1}$ and the number of interior points is trivially bounded by $n$, and by our assumptions $\frac{n}{m_n + 1} \rightarrow 0$.
\end{proof}

\begin{proof}[Proof of proposition \ref{prop:driving_process}]
    Note that the only way to increase the number of open points is to query a point which is not open, and for that point to have label 0. In this case we increase the number of open points by at most 2. This is the event $\{ Z_i = 1,Y_i = 0 \}$. Conversely if we query an open point and it's label is 1 then we decrease the number of open points by at least 1. This is the event $\{ Z_i = 0,Y_i = 1 \}$. And even when neither of these happens the number of open points can still decrease, but cannot increase. Thus we have that $U'_i = \max\{supp(U_i | Y_i = y, Z_i = z) \}$. 
\end{proof}

\begin{proof}[Proof of proposition \ref{prop:hit_0_io}]
    We will prove by induction that $\tau_j < \infty \: \forall \: j$ with probability 1. Note for our base case that $\tau_1 = 0 < \infty$. Now assume $\tau_{j-1} = i_0 < \infty$. If $U_{i_0+1} = 0$ (which can happen if for example our new data point has label 1) then $\tau_j = i_0+1 < \infty$. Now assume $U_{i_0+1} > 0$. Thus we know that $O_{i_0+1} > 0$, and will remain above 0 until $\tau_j$ giving us that $\sum\limits_{i = i_0+2}^{\tau_j} U_i \leq \sum\limits_{i = i_0+2}^{\tau_j} U'_i$. But for some $i_1$ $E[U'_i] < 0 \: \forall \: i \geq i_1$. Therefore by SLLN $\sum\limits_{i = i_0}^{\infty} U'_i \stackrel{a.s.}{\rightarrow} -\infty$ and so there exists some $T < \infty$ s.t. $\sum\limits_{i = i_0}^{\infty} U'_i \leq 0$ with probability 1. Therefore $\tau_j \leq T < \infty$ with probability 1, and so $P(O_i = 0 \: i.o.) = 1$.
\end{proof}

\begin{proof}[Proof of proposition \ref{prop:err_before_rand_one}]
\begin{enumerate}[i]
\item
\begin{equation*}
\begin{aligned}
    &P(T_{1,i_0} < T_{2,i_0}) =\\
    &\sum\limits_{t=1}^{\infty} P(T_{1,i_0} = t)P(T_{1,i_0} < T_{2,i_0} | T_{1,i_0} = t)\\
    &= \sum\limits_{t=1}^{\infty} p_{i_0+t} \prod\limits_{j=1}^{t-1} (1 - p_{i_0+j}) [(1-\eta)^{t-1}]\\
    &\leq p_{i_0} \sum\limits_{t=1}^{\infty}  [(1-\eta)^{t-1}] = p_{i_0} \frac{1}{\eta}\\
\end{aligned}
\end{equation*}
\item
\begin{equation*}
\begin{aligned}
    &P(T_{1,i_0} = T_{2,i_0}) = \sum\limits_{t=1}^{\infty} P(T_{1,i_0} = t) P(T_{2,i_0} = t)\\
    &= \sum\limits_{t=1}^{\infty} p_{i_0+t} \prod\limits_{j=1}^{t-1} (1 - p_{i_0+j}) [P(T_{2,i_0} = i_0+t)]\\
    &\leq p_{i_0} \sum\limits_{t=1}^{\infty}  [P(T_{2,i_0} = i_0+t)] = p_{i_0}\\
\end{aligned}
\end{equation*}
\end{enumerate}
\end{proof}

\begin{proof}[Proof of corollary \ref{cor:err_before_rand_k}]
    The three inequality relationships come straight from the independence of our random variables $Y_i, Z_i$. The final statement can be shown by induction. It is clearly true for the case $a=1, b=1$. Assume true for all $a \leq a_0 - 1, b \leq b_0$. 
    
    \begin{equation*}
    \begin{aligned}
        p_{i_0}^{(a_0,b_0)} \leq& p_{i_0}^{(1,1)} \times 3^{a_0-1+b_0} (p_{i_0}^{(1,1)})^{a_0-1} + \\
        &\eta p_{i_0}^{(1,1)} \times 3^{a_0-1 + b_0 - 1} (p_{i_0}^{(1,1)})^{a_0-1} +\\
        &3^{a_0 + b_0 - 1} (p_{i_0}^{(1,1)})^{a_0}\\
        \leq& 3^{a_0+b_0} (p_{i_0}^{(1,1)})^{a_0}
    \end{aligned}
    \end{equation*}
    And finally note that in the above there is symmetry between the roles of $a$ and $b$ so the same calculations show that if it's true for all $a \leq a_0, b \leq b_0 - 1$ then it's true for $a \leq a_0, b \leq b_0$.
    
    % Now assume true for all $a \leq a_0, b \leq b_0 - 1$.
    
    % \begin{equation*}
    % \begin{aligned}
    %     p_{i_0}^{(a_0,b_0)} \leq& p_{i_0}^{(1,1)} \times 3^{a_0-1+b_0} (p_{i_0}^{(1,1)})^{a_0-1} + \\
    %     &\eta p_{i_0}^{(1,1)} \times 3^{a_0-1 + b_0 - 1} (p_{i_0}^{(1,1)})^{a_0-1} + 3^{a_0 + b_0 - 1} (p_{i_0}^{(1,1)})^{a_0}\\
    %     \leq& 3^{a_0+b_0} (p_{i_0}^{(1,1)})^{a_0}
    % \end{aligned}
    % \end{equation*}
\end{proof}

\begin{proof}[Proof of lemma \ref{lemma:few_open}]
    Let $E_j$ be the event that during the $j^{th}$ cycle we have more than 4 open points. So if the $j^{th}$ cycle starts at time $\tau_j$ then $E_j = \{\max\limits_{\tau{j} \leq j \leq \tau_{j+1}}O_j > 4 \}$. Also $\forall \: t \in [\tau{j} \leq j \leq \tau_{j+1}], \{ O_t > 4 \} \subset E_j$. Note that $P(E_j) \leq p_{\tau_{j}}^{(2,4)} \leq c p_{\tau_{j}}^2 \leq c p_{j-1}^2$ since each cycle must have length at least 1. Thus if we hit $n$ during the $j^{th}$ cycle then $P(O_n > 4) \leq c p_j^2 \rightarrow 0$ as $j \rightarrow \infty$. And by proposition \ref{prop:hit_0_io} we have that if $j_0$ is the cycle we are in at time $n$ then $j_0 \rightarrow \infty \: a.s.$ as $n \rightarrow \infty$. Thus $P(E_{j_0}) \rightarrow 0$.
\end{proof}

\begin{proof}[Proof of proposition \ref{prop:open_point}]
    Fix $\epsilon > 0$ and $\delta < \frac{\epsilon}{2}$. Define two events:
    \begin{enumerate}
        \item $\Omega_1 = $ \{we return to 0 infinitely often\}
        \item $\Omega_2 = $ \{$\forall \: i \: X_{(i+1)} - X_{(i)} \leq \delta$ and $X_{(1)}, 1 - X_{(n)} \leq \delta$\}
    \end{enumerate}
    then $\{d_n > \epsilon \: \forall \: n\} \subset (\Omega_1 \cap \Omega_2)^c$. This is because returning to 0 infinitely often means that infinitely often we act according to $A^{\dagger}$ when the number of open points is 0. This action samples the unlabelled point which is furthest from any labelled point. We will show that just these actions are enough to prevent $d_n \geq \epsilon \: \forall \: n$ when (2) is also true. We will also ignore the fact that our labelled intervals take up length as this length is negligible and only forces the \textit{empty interval} (the interval of consecutive unlabelled points) to be smaller. 
    
    By (2) if the empty interval containing the unlabelled point which is furthest from any labelled point is of size $l$ then the point which is newly labelled must be within $\frac{\delta}{2}$ of the center of the interval, and so the maximum size of the two new empty intervals created is $\frac{l + \delta}{2}$. If $l \geq \epsilon$ then we get that the new empty intervals have length $\leq \frac{3}{4}l$, so we're guaranteed to produce empty intervals of length no more than $\frac{3}{4}$ of the original intervals length. Additionally since $\epsilon > 2\delta$ there are no empty intervals which cannot be cut to size smaller than $\epsilon$ due to there not being two consecutive points with distance greater than $\epsilon$. So any interval of finite size $> \epsilon$ can be split into intervals all of size less than $\epsilon$ in a finite number of cuts. Thus if at any time $t$ we have $N < \infty$ empty intervals of size $> \epsilon$ (which must be the case since the sum of our interval lengths is bounded by 1) they will all be reduced to intervals of size $< \epsilon$ in a finite number of cuts.
    
    % If $x$ is one of the labelled points around this \textit{empty interval}, then by (2) there must be an unlabelled point $x'$ within $\frac{\delta}{2}$ of the center of this empty interval. Thus $|x - x'| \geq \frac{\epsilon - \delta}{2}$, and every time an empty interval is cut the size of the two new empty intervals must be $\leq \frac{\epsilon + \delta}{2}$. So for $d_n > \epsilon \: \forall \: n$ there must be an infinite number of points further than $\frac{\epsilon - \delta}{2}$ from any interval, which is impossible since the length of our space is 1. 
    
    By proposition \ref{prop:hit_0_io} $P(\Omega_1) = 1$. By Glivenko-Cantelli $P(\Omega_2) = 1$, since otherwise $\exists \: x \: s.t. F_n(x) = F_n(x + \delta) \: \forall \: n$, where $F_n(x)$ is the usual empirical cdf. But $F(x) \neq F(x + \delta)$ and so Glivenko-Cantelli would be violated, which happens with probability 0. Therefore with probability 1 we cannot have that $d_n > \epsilon \: \forall \: n$ and so $d_n \stackrel{a.s.}{\rightarrow} 0$.
\end{proof}

\begin{proof}[Proof of corollary \ref{cor:open_point}]
    The coverage of each labelled point adjacent to an open point is half the distance to the next interval. However by \ref{prop:open_point} this distance $\stackrel{a.s.}{\rightarrow} 0$. 
\end{proof}

\begin{proof}[Proof of theorem \ref{thrm:1nn}.]

    Let $\mathcal{I}_n$ be the set of all interior points and let $\mathcal{A}_n$ be the set of all labelled points adjacent to open points. $E[\mathbf{1}_{f(X, S_n) = 1}] = E \Big{[} \sum\limits_{x \in S_n}  I(x, S_n) \mathbf{1}_{L(x) = 1}\Big{]} = E \Big{[} \sum\limits_{x \in \mathcal{I}_n}  I(x, S_n) \mathbf{1}_{L(x) = 1} + \sum\limits_{x \in \mathcal{A}_n}  I(x, S_n) \mathbf{1}_{L(x) = 1} + \sum\limits_{x \in S_n \setminus \mathcal{I}_n \cup \mathcal{A}_n}  I(x, S_n) \mathbf{1}_{L(x) = 1} \Big{]}$. We know that all points with label 0 are either in $\mathcal{I}_n$ or $\mathcal{A}_n$. By corollary \ref{cor:interior} $E\sum\limits_{x \in \mathcal{I}_n}  I(x, S_n) \rightarrow 0$, and by corollary \ref{cor:open_point} $E\sum\limits_{x \in \mathcal{A}_n}  I(x, S_n) \rightarrow 0$. Thus since $E\sum\limits_{x \in S_n}  I(x, S_n) = 1$ we have that $E\sum\limits_{x \in S_n \setminus \mathcal{I}_n \cup \mathcal{A}_n}  I(x, S_n) \rightarrow 1$, and since $\mathbf{1}_{L(x) = 1} = 1 \: \forall \: x \in S_n \setminus \mathcal{I}_n \cup \mathcal{A}_n$ we have that $E[\mathbf{1}_{f(X, S_n) = 1}] \rightarrow 1$, and so $E[\mathbf{1}_{f(X, S_n) \neq Y}] \rightarrow 1-\eta$.
\end{proof}

\begin{remark}
    It is clear that a similar result for regression (with squared loss) could be obtained using the same idea, with $f^*(X) = E[Y | X ] = 0, Y_i = \epsilon_i$ (where $\epsilon_i$ is our \textit{iid} $E[\epsilon]=0$ noise) by using the above algorithm. Let a point have a pseudo-label of 0 if $|Y| \leq c \in |supp(\epsilon)|$ and 1 otherwise, and run the above algorithm on the pseudo-labels. You would again get intervals of low value points enclosed by high value points and could get MSE $\geq c^2$.
\end{remark}

\begin{proof}[Proof of proposition \ref{prop:close_before_rand_k}]
\begin{enumerate}[i]
\item
\begin{equation*}
\begin{aligned}
    &P(T_{1,{i_0}} < T_{2,{i_0}}) =\\ 
    &\sum\limits_{t=1}^{\infty} P(T_{1,{i_0}} = t) P(T_{1,{i_0}} < T_{2,{i_0}} | T_{1,{i_0}} = t)\\
    &P(T_{1,{i_0}} = t) \leq \frac{1}{{i_0}+t}\\
    &\text{By Markov}\\
    &P(T_{2,{i_0}} > t) \leq \frac{E[T_{2,{i_0}}]}{t+1} \leq \frac{E[T_{2,{i_0}}]}{t}\\
    &E[T_{2,{i_0}}] = \frac{1}{1-\eta} \big{(} \frac{1}{\eta ^{k'}} - 1 \big{)}\\
    &\text{By AM-GM inequality}\\
    &P(T_{1,{i_0}} < T_{2,{i_0}}) \leq \sum\limits_{t = 1}^{\infty} \frac{1}{{i_0}+t} \frac{1}{t} \frac{1}{1-\eta} \big{(} \frac{1}{\eta ^{k'}} - 1 \big{)}\\ 
    &\leq \frac{1}{2 \sqrt{{i_0}}} \frac{1}{1-\eta} \big{(} \frac{1}{\eta ^{k'}} - 1 \big{)} \sum\limits_{t = 1}^{\infty} \frac{1}{t^{\frac{3}{2}}}\\
    &=\frac{c}{2 \sqrt{{i_0}}} \frac{1}{1-\eta} \big{(} \frac{1}{\eta ^{k'}} - 1 \big{)}
\end{aligned}
\end{equation*}
\item Proof is same as for proposition \ref{prop:err_before_rand_one}
\end{enumerate}
\end{proof}

\begin{proof}[Proof of lemma \ref{lemma:under_6_as}]
    Let $E_j$ be the event that during the $j^{th}$ cycle we have more than 6 open points. So if the $j^{th}$ cycle starts at time $\tau_j$ then $E_j = \{\max\limits_{\tau{j} \leq j \leq \tau_{j+1}}O_k > 6 \}$. Also $\forall \: t \in [\tau{j} \leq j \leq \tau_{j+1}], \{ O_t > 6 \} \subset E_j$. Note that $P(E_j) \leq p_{\tau_{j}}^{(3,6)} \leq (c p_{\tau_{j}})^3 \leq (c p_{j-1})^3 = c^3\frac{1}{j^{\frac{3}{2}}}$. By proposition \ref{prop:hit_0_io} we have that if $j_0$ is the cycle we are in at time $n$ then $j_0 \rightarrow \infty \: a.s.$ as $n \rightarrow \infty$. And by Borel-Cantelli we have that $P(E_{j} \: i.o.) = 0$. 
\end{proof}

\subsection{SUFFICIENCY FOR BOUNDED SUPPORT ESTIMATORS}

\begin{proof}[Proof of Theorem \ref{thrm:bounded_sup_consistency}]
    
    For convenience of notation, we will let $Y_i \in \{1, -1\}$, using the usual transformation from our current $Y_i \in \{0,1\}$ setting. Under this transformation, and by the assumptions on the structure of our $W_{ni}(x, S_n(X))$,
    
    \begin{equation*}
    \begin{aligned}
        f_n(x, S_n) &= sign(\sum W_{ni}(x, S_n(X))Y_i)\\
        &= sign(\sum w_n(x, X_i)Y_i)
    \end{aligned}
    \end{equation*}
    
    Therefore for consistency we want to show that $P( f^*(X) f_n(X, S_n) = -1 ) \rightarrow 0$. For $x$ fixed this occurs iff $\sum w_n(x, X_i)Y_i f^*(x) < 0$. Define $\gamma_n = \{x \in \mathcal{X} : \sup\limits_{z \in supp_n(x)} |f(z) - f(x)| \leq \frac{|0.5 - f(x)|}{2} \}$. By the assumption that $diam(supp_n(x)) \rightarrow 0$ and Regularity Condition \ref{cond:conditional_cont}, $P_X(\gamma_n) \rightarrow 1$ , and so for some $\epsilon > 0$, for $n$ sufficiently large 
    
    \begin{equation*}
    \begin{aligned}
        &P( f^*(X) f_n(X, S_n) = -1 )\\
        &\leq P( f^*(X) f_n(X, S_n) = -1 | X \in \gamma_n ) + \epsilon.
    \end{aligned}
    \end{equation*}
    
    Also define the following:
    
    \begin{equation*}
    \begin{aligned}
        &S_n(x) = \sum w_n(x, X_i) Y_i f^*(x) = S_n^{(R)}(x) + S_n^{(A)}(x)\\
        &S_n^{(R)}(x) = \sum w_n(x, X_i) Z_i Y_i f^*(x)\\
        &S_n^{(A)}(x) = \sum w_n(x, X_i)(1- Z_i) Y_i f^*(x)\\
    \end{aligned}
    \end{equation*}
    
    We want to show that $P(S_n^{(R)}(X) \leq M | X \in \gamma_n) \rightarrow 0 \: \forall \: M < \infty$. To do this we will lower bound $S_n^{(R)}(x)$ by a sum which is easier to analyze, and prove that sum diverges in probability if $x \in \gamma_n$. Let 
    
    \begin{equation*}
    \begin{aligned}
        &\tilde{S}_n^{(R)}(x) = \sum w_n(x, X_i) Z_i \tilde{Y}_i(x, X_i, Y_i) f^*(x)\\
        &\tilde{Y}_i(x, X_i, Y_i) = Y_i\mathbf{1}_{Y_i \neq f^*(x)} +  Y'_i(x, X_i)\mathbf{1}_{Y_i = f^*(x)}\\
        &Y'_n(x, X_i) \in \{1, -1 \}\\
        &P(Y'_n(x, X_i) = f^*(x)) =\\
        &\begin{cases}
        &\frac{\inf\limits_{z \in supp_n(x)} f(z)}{f(X_i)} \quad \text{ if } f^*(x) = 1, X_i \in supp_n(x) \\
        &\frac{\inf\limits_{z \in supp_n(x)}1-f(z)}{1-f(X_i)} \quad \text{ if } f^*(x) = -1, X_i \in supp_n(x) \\
        &0 \quad \text{otherwise}\\
        \end{cases}\\
        &\text{Where the randomness in $Y_i'$ is independent of everything}
    \end{aligned}
    \end{equation*}
    
    So by construction we have that
    $P(\tilde{Y}_i = f^*(x) | X_i \in supp_n(x))$ 
    $= \inf\limits_{z \in supp_n(x)} P(\tilde{Y}_i = f^*(z)|X_i \in supp_n(x))$,
    $\tilde{Y}_i | X_i \in supp_n(x) \perp X_i$ and $Y_i f^*(x) \geq \tilde{Y}_i f^*(x)$. Since $x \in \gamma_n, P(\tilde{Y}_i f^*(x) = 1 | X_i \in supp_n(x)) > \frac{1}{2} + \frac{|0.5 - f(x)|}{2}$. By Condition \ref{property:bounded_sup} we have that $\sum\limits_{i=1}^n \mathbf{1}_{X_i \in supp_n(X), Z_i = 1} \CinP \infty$ and $w_n(x, X_i) | Z_i = 1 \stackrel{d}{=} w_n(x, X_j) | Z_j = 1$, which gives us that $P(\tilde{S}_n^{(R)}(X) \leq M | X \in \gamma_n) \rightarrow 0 \: \forall \: M < \infty$ which in turn gives us that $P(S_n^{(R)}(X) \leq M | X \in \gamma_n) \rightarrow 0 \: \forall \: M < \infty$. And since $\epsilon$ was arbitrary this gives us $P(S_n^{(R)}(X) \leq M) \rightarrow 0 \: \forall \: M < \infty$.
    
    Now in order for $S_n(x) < 0$ we need that $S_n^{(A)}(x) \rightarrow -\infty$. By defining $\tilde{S}_n^{(A)}(x)$ similarly, the same argument shows that this cannot happen. Since $w_n(x,y) \leq K$ we would require an infinite number of active samples in $supp_n(x)$. For each of these we would have $ P({Y}_i f^*(x) = 1 | X_i \in supp_n(x)) > \frac{1}{2} + \frac{|0.5 - f(x)|}{2}$, and so even though we can stop as soon as we are smaller than $M$, $P(\tilde{S}_n^{(A)}(x) \leq M) \rightarrow 0$ as $M \rightarrow -\infty$. Therefore $P(S_n(X) > 0) \rightarrow 1$ and $P( f^*(X) f_n(X, S_n) = -1 ) \rightarrow 0$. 
    
\end{proof}

%\begin{figure}[h!]
	%\caption{}
	%\centering
	%\begin{tabular}{cc}
		%\includegraphics[width=0.99\textwidth,height=0.25\textheight,clip = true]{}\\
	%\end{tabular}
	%\label{fig:diff}
%\end{figure}
%
%
%
%\begin{equation*}
%\begin{aligned}
%\end{aligned}
%\end{equation*}
%
%
%
%\begin{lstlisting}[language=Python,basicstyle=\tiny]
%
%
%
%
%\end{lstlisting}

% \vfill
% \FloatBarrier \pagebreak
% \nocite{*}

\end{document}